\documentclass{article}

\PassOptionsToPackage{numbers, compress}{natbib}

\usepackage[final]{neurips_2024}

\usepackage[utf8]{inputenc} 
\usepackage[T1]{fontenc}    
\usepackage{hyperref}       
\usepackage{url}            
\usepackage{booktabs}       
\usepackage{amsfonts}       
\usepackage{nicefrac}       
\usepackage{microtype}      
\usepackage{xcolor}         
\usepackage{amsmath}
\usepackage{graphicx}
\usepackage{multirow}
\usepackage{adjustbox}

\title{GeoLRM: Geometry-Aware Large Reconstruction Model for High-Quality 3D Gaussian Generation}

\author{%
  Chubin Zhang$^{1,3}$ \quad Hongliang Song$^{3}$ \quad Yi Wei$^{2}$\\
  \textbf{Yu Chen}$^{3}$ \quad \textbf{Jiwen Lu}$^{2}$ \quad \textbf{Yansong Tang}$^{1, \ddagger}$\\
  $^1$Tsinghua Shenzhen International Graduate School, Tsinghua University\\
  $^2$Department of Automation, Tsinghua University\\
  $^3$Alibaba Group\\
  \texttt{\{zcb24, y-wei19\}@mails.tsinghua.edu.cn},\\
  \texttt{\{hongliang.shl, chenyu.cheny\}@alibaba-inc.com},\\
  \texttt{lujiwen@tsinghua.edu.cn},  \texttt{tang.yansong@sz.tsinghua.edu.cn}.\\
  {\small $^{\ddagger}$ corresponding author}
}

\begin{document}

\maketitle

\begin{abstract}
  In this work, we introduce the Geometry-Aware Large Reconstruction Model (GeoLRM), an approach which can predict high-quality assets with 512k Gaussians and 21 input images in only 11 GB GPU memory. Previous works neglect the inherent sparsity of 3D structure and do not utilize explicit geometric relationships between 3D and 2D images. This limits these methods to a low-resolution representation and makes it difficult to scale up to the dense views for better quality. GeoLRM tackles these issues by incorporating a novel 3D-aware transformer structure that directly processes 3D points and uses deformable cross-attention mechanisms to effectively integrate image features into 3D representations. We implement this solution through a two-stage pipeline: initially, a lightweight proposal network generates a sparse set of 3D anchor points from the posed image inputs; subsequently, a specialized reconstruction transformer refines the geometry and retrieves textural details. Extensive experimental results demonstrate that GeoLRM significantly outperforms existing models, especially for dense view inputs. We also demonstrate the practical applicability of our model with 3D generation tasks, showcasing its versatility and potential for broader adoption in real-world applications. The project page: \url{https://linshan-bin.github.io/GeoLRM/}.
\end{abstract}

\section{Introduction}
\label{intro}

In fields ranging from robotics to virtual reality, the quality and diversity of 3D assets can dramatically influence both user experience and system efficiency. Historically, the creation of these assets has been a labour-intensive process, demanding the skills of expert artists and developers. While recent years have witnessed groundbreaking advancements in 2D image generation technologies, such as diffusion models~\cite{stablediff, imagen, dalle} which iteratively refine images, their adaptation to 3D asset creation remains challenging. Directly applying diffusion models to 3D generation~\cite{shape, pointe} is less than satisfactory, primarily due to a dearth of large-scale and high-quality data. DreamFusion~\cite{dreamfusion} innovatively optimize a 3D representation~\cite{mipnerf} by distilling the score of image distribution from pre-trained image diffusion models~\cite{stablediff, imagen}. However, this approach lacks a deep integration of 3D-specific knowledge, such as geometric consistency and spatial coherence, leading to significant issues such as the multi-head problem and the inconsistent 3D structure. Additionally, these methods require extensive per-scene optimizations, which severely limits their practical applications.

The introduction of the comprehensive 3D dataset Objaverse~\cite{objaverse, objaversexl} brings significant advancements for this field. Utilizing this dataset, researchers have fine-tuned 2D diffusion models to produce images consistent with 3D structures~\cite{zero123, zero123++, mvdream}. Moreover, recent innovations~\cite{xu2024grm, wang2024crm, tang2024lgm, xu2024instantmesh, wei2024meshlrm} have combined these 3D-aware models with large reconstruction models (LRMs)~\cite{hong2023lrm} to achieve rapid and accurate 3D image generation. These methods typically employ large transformers or UNet models that convert sparse-view images into 3D representations in a single forward step.
While they excel in speed and maintaining 3D consistency, they confront two primary limitations. Firstly, previous works utilize triplanes~\cite{hong2023lrm, xu2024instantmesh, wang2024crm} to represent the 3D models, wasting lots of features in regions devoid of actual content and involving dense computations during rendering. This \textit{violates the sparse nature of 3D} as our analysis shows that the visible portions of the 3D models in the Objaverse dataset constitute only about 5\% of the overall spatial volume. Though Gaussian-based methods~\cite{tang2024lgm, xu2024grm, wei2024meshlrm} may use pixel-aligned Gaussians for better efficiency, this representation is incapable of recovering the unseen area and thus heavily relies on the input images. Secondly, previous works tend to \textit{overlook the explicit geometric relationships between 3D and 2D images}, which results in ineffective processing. The tri-plane or pixel-aligned Gaussian tokens do not correspond to a specific space in 3D, thus being unable to utilize the projection relationship between 3D points and images. In other words, they conduct dense attention between the 3D queries and the image keys. This leads to the fact that these methods tend to reconstruct 3D with sparse view inputs but cannot achieve better performance with denser inputs.

To address these challenges, we introduce the geometry-aware large reconstruction model (GeoLRM) for 3D Gaussian generation. Our method centres on a 3D-aware reconstruction transformer that eschews conventional representations like triplanes or pixel-aligned Gaussians in favour of a direct interaction within the 3D space. However, directly generating 3D Gaussians in the whole 3D space requires huge memory costs. To this end, we first propose a specialized proposal network to predict an occupancy grid from input images. Only the occupied voxels will be further processed to generate 3D Gaussian features. The proposed transformer replaces the dense cross attention with deformable cross attention~\cite{deform_detr}. By projecting the input 3D tokens onto the corresponding image planes, these tokens only focus on the most relevant features, which greatly improves the effectiveness.

\begin{figure}[t]
  \centering
  \includegraphics[width=0.9\textwidth]{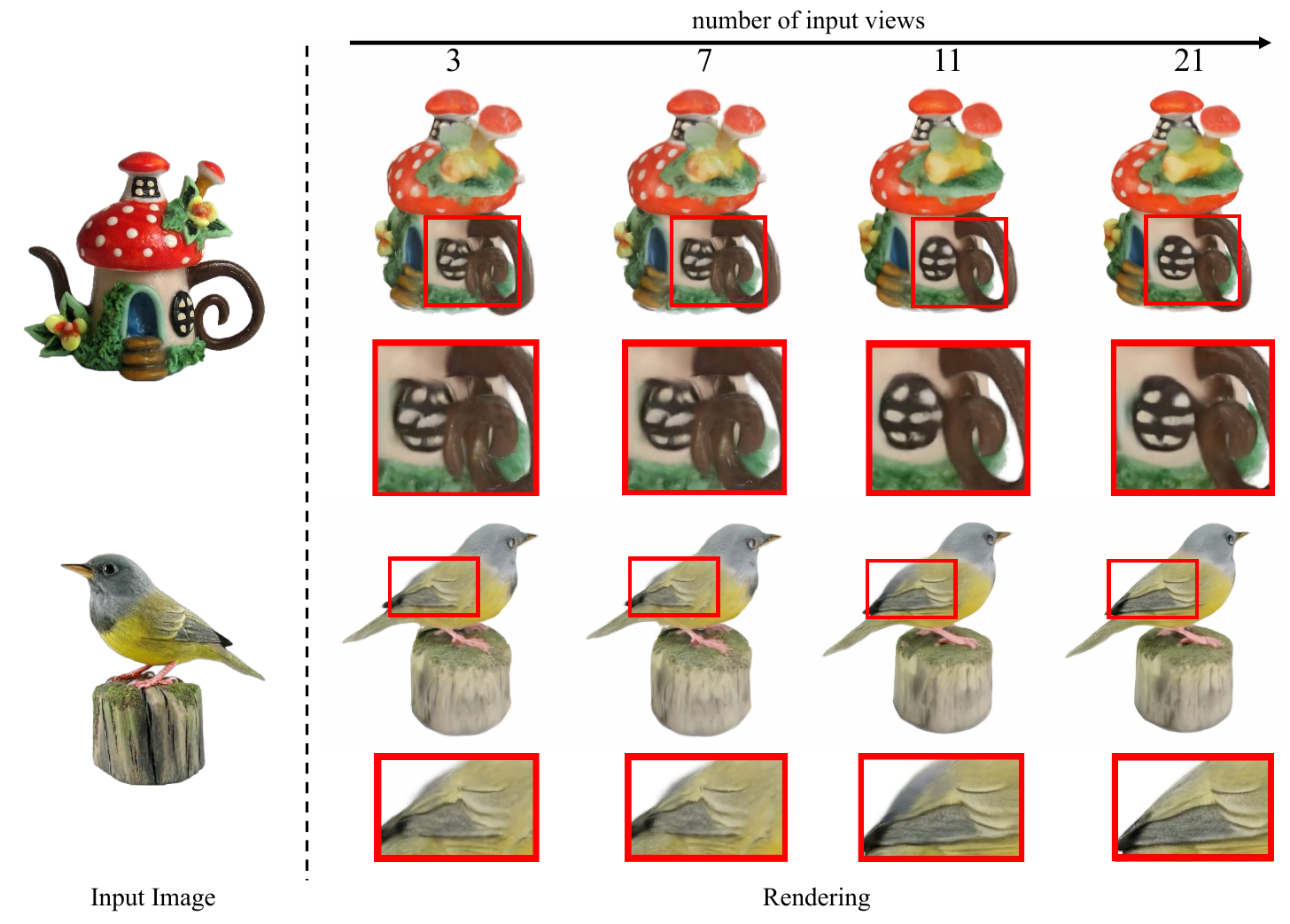}
  \caption{Image to 3D using GeoLRM. Initially, a 3D-aware diffusion model, specifically SV3D~\cite{sv3d}, transforms an input image into multiple views. Subsequently, these views are processed by our GeoLRM to generate detailed 3D assets. \textbf{Unlike other LRM-based approaches, GeoLRM notably improves as the number of input views increases.}}
\label{fig:teaser}
\vspace{-15pt}
\end{figure}

We trained our GeoLRM on the Objaverse dataset rendered by~\cite{richdreamer} and tested it on the Google Scanned Objects~\cite{gso}. By integrating geometric principles, our model not only outperforms existing methods with the same number of inputs but also makes it possible to work with denser image inputs. Significantly, the model efficiently handles up to 21 images (even more if necessary), yielding superior 3D models in comparison to those generated from fewer images. Leveraging this capability, we integrated GeoLRM with SV3D~\cite{sv3d} for high-quality 3D model generation.

In summary, our contributions are as follows:

\begin{itemize}
\item We introduce a two-stage pipeline that leverages the sparse nature of 3D data, resulting in a sparse 3DGS token representation suitable for extension to high resolution.
\item We fully exploit the projection relationship between 3D points and 2D images, significantly reducing the space complexity of attention mechanisms in LRMs, thus enabling denser image input configurations.
\item To the best of our knowledge, GeoLRM is the first to process dense inputs using LRM, potentially paving the way for integrating video generation models into 3D AIGC applications.
\end{itemize}

\section{Related Work}

\subsection{Optimization-based 3D reconstruction}
3D reconstruction from multi-view images has been extensively studied in computer vision for decades. While traditional methods like SfM~\cite{sfm1, sfm2, sfm3} and MVS~\cite{mvs1, mvs2} provide basic reconstruction and calibration, they lack robustness and expressiveness. Recent advancements leverage learning-based methods for better performance. Among these methods, NeRF~\cite{nerf} stands out for its capability of capturing high-frequency details. Following works~\cite{mipnerf, nerf++, mip360, instantngp, plenoxels, chen2022tensorf, sun2021direct, zipnerf} further improve its performance and speed. Though NeRF has made a great improvement, the need to query tons of points during the rendering process makes it hard for real-time applications. 3D Gaussians~\cite{3dgs} solves this problem by explicitly expressing a scene with 3D Gaussians and utilizing an efficient rasterization pipeline. These methods involve a per-scene optimization process and require dense multi-view images for a good reconstruction.

\subsection{Large Reconstruction Model}
Different from optimization-based 3D reconstruction methods, large reconstruction models~\cite{hong2023lrm,li2023instant3d,tang2024lgm,xu2024grm,wei2024meshlrm,zhang2024gslrm,wang2023pf,wang2024crm} are able to reconstruct 3D shapes in a feed-forward way. As the pioneer work of this area, the LRM~\cite{hong2023lrm} illustrates that the transformer backbone can effectively leverage the power of large-scale datasets and translate image tokens into implicit 3D triplanes under multi-view supervision. Beyond LRM, Instant3D ~\cite{li2023instant3d} improves reconstruction quality with sparse-view inputs. It employs a two-stage paradigm, which first generates four views with the diffusion model and then regresses NeRF~\cite{nerf} from generated multi-view images. Instead of NeRF, InstantMesh~\cite{xu2024instantmesh} utilizes mesh representation to reconstruct 3D objects, which adopts a differentiable iso-surface extraction module. However, many works~\cite{tang2024lgm,zhang2024gslrm,xu2024grm,xu2024agg} choose 3D Gaussians~\cite{3dgs} as the outputs. GRM~\cite{xu2024grm} proposes a transformer network to translate pixels to the set of pixel-aligned 3D Gaussians while LGM~\cite{tang2024lgm} uses an asymmetric UNet to predict and fuse 3D Gaussians. Compared with these methods, our GeoLRM projects multi-view features to the 3D space with cross-view attention mechanisms, which explicitly explores geometric knowledge.

\subsection{3D generation}
Early methods~\cite{chan2022efficient,chan2021pi,gao2022get3d,nguyen2019hologan,skorokhodov2022epigraf,xu20223d,niemeyer2021giraffe} in 3D generation area utilize 3D GANs to generate 3D-aware contents. Although some methods ~\cite{melas2023pc2,melas2023pc2,zhou20213d,liu2023meshdiffusion,chou2023diffusion,shim2023diffusion,zhang20233dshape2vecset} replace 3D GANs with 3D diffusion models for high-quality generation, their generalization ability is bounded by the limited training data. Recently, proposed in DreamFusion~\cite{dreamfusion}, score distillation sampling (SDS) requires no 3D data and is able to leverage the great power of 2D text-to-image diffusion models~\cite{imagen,stablediff,dalle}. Specifically, it optimizes a randomly-initialized 3D model and diffuses the render images with a pretrained diffusion model. As the follow-up works~\cite{prolificdreamer,chen2023fantasia3d,lin2023magic3d,wang2023score,tang2023dreamgaussian,yi2023gaussiandreamer,liu2023sherpa3d,yu2023text,liang2023luciddreamer,li2023sweetdreamer,richdreamer}, many methods have been proposed to accelerate the optimization process or improve 3D generation quality. Different with SDS-based methods, Zero-1-to-3~\cite{zero123} fine-tunes the 2D diffusion models on a large-scale synthetic dataset to change the camera viewpoint of a given image.  Similar to Zero-1-to-3, many other works~\cite{zero123++,sv3d,mvdream,dmv3d,liu2023syncdreamer,weng2023consistent123,long2023wonder3d,woo2023harmonyview} aim to synthesize multi-view consistent images. Our method can reconstruct 3D contents based on these synthesis multi-view images.   

\section{Methodology}

\begin{figure}[t]
  \centering
  \includegraphics[width=1.0\textwidth]{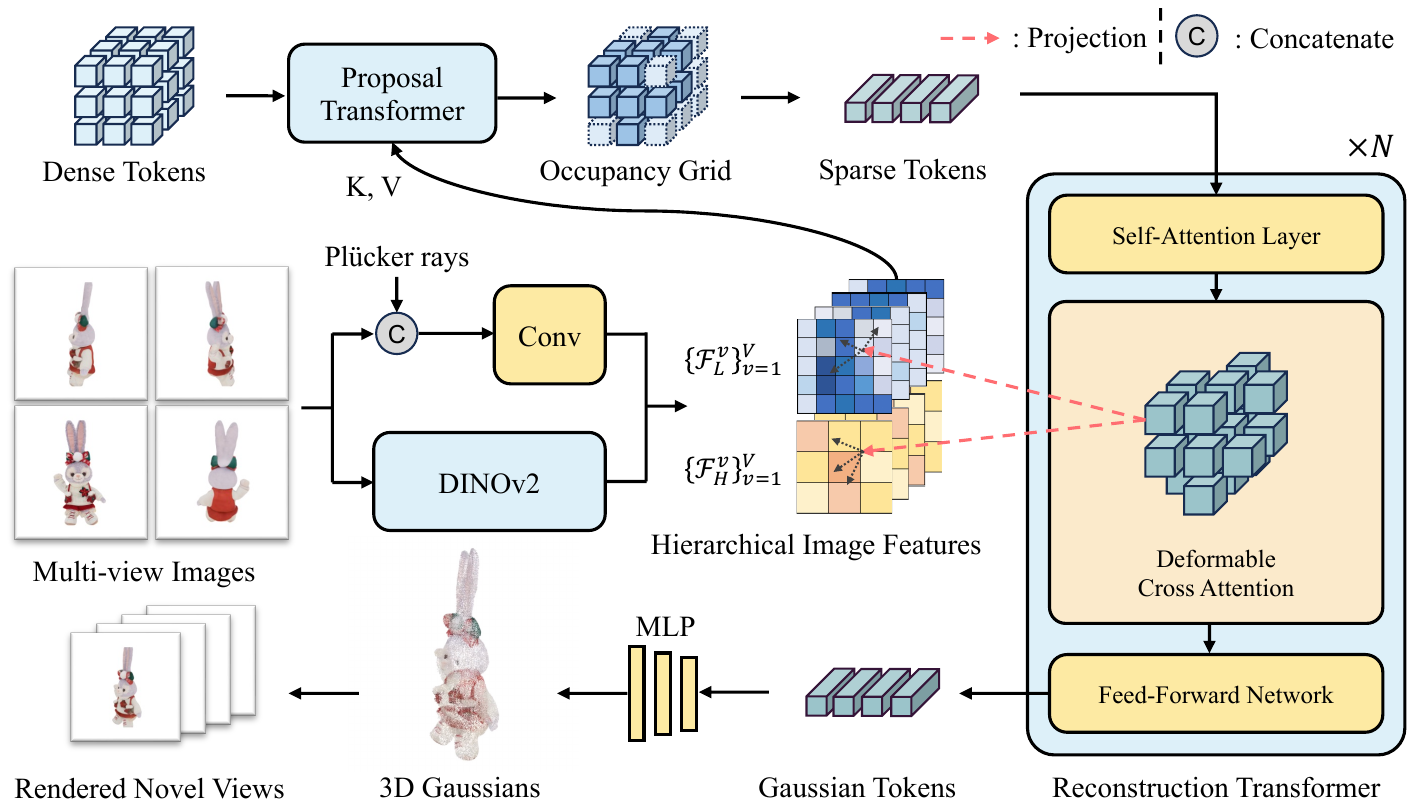}
  \caption{\textbf{Pipeline of the proposed GeoLRM}, a geometry-powered method for efficient image to 3D reconstruction. The process begins with the transformation of dense tokens into an occupancy grid via a Proposal Transformer, which captures spatial occupancy from hierarchical image features extracted using a combination of a convolutional layer and DINOv2~\cite{dinov2}. Sparse tokens representing occupied voxels are further processed through a Reconstruction Transformer that employs self-attention and deformable cross-attention mechanisms to refine geometry and retrieve texture details with 3D to 2D projection. Finally, the refined 3D tokens are converted into 3D Gaussians for real-time rendering.}
\label{fig:pipeline}
\vspace{-5pt}
\end{figure}

\subsection{Overview}
\label{sec:method_ov}

Figure~\ref{fig:pipeline} illustrates the pipeline of our proposed method. Our approach takes a set of images $\{{I}^{i}\}_{i=1}^{N}$ with their corresponding intrinsic $\{{K}^{i}\}_{i=1}^{N}$ and extrinsic $\{{T}^{i}\}_{i=1}^{N}$ as input. Initially, we encode input images into hierarchical image features and predict an occupancy grid with a proposal transformer. Each occupied voxel within this grid is considered a 3D anchor point. These 3D anchor points are then processed by a reconstruction transformer, refining their geometry and retrieving textural details. The proposal and reconstruction transformers share the same model architecture, which is further discussed in Section~\ref{sec:method_ma}. The outputs of the reconstruction transformer are decoded into Gaussian features with a shallow MLP for rendering. Loss functions are described in Section~\ref{sec:method_to}.

\subsection{Model Architecture}
\label{sec:method_ma}
Our model architecture features a hierarchical image encoder for extracting high and low-level image feature maps along with a geometry-aware transformer for lifting 2D features into 3D representations.

\textbf{Hierarchical Image Encoder}\quad Our method integrates both high and low-level features to enhance model performance. For high-level features, we utilize DINOv2~\cite{dinov2}, which excels in single-image 3D tasks~\cite{banani2024probing}. To capture low-level features, we combine Plücker ray embeddings and RGB values. The Plücker ray parameterizes each ray corresponding to a pixel by $\textbf{r}=(\textbf{d}, \textbf{o}\times\textbf{d})$, with $\textbf{d}$ representing the ray's direction and $\textbf{o}$ its origin \cite{lightfield, dmv3d}. These embeddings, denoted as $R^v$ for each image $I^v$, are concatenated with the RGB values of the image. This combined data is then integrated through a convolution layer. The encoding processes are succinctly described by the equations:
\begin{align}
    \mathcal{F}_H^v &= \text{DINOv2}(I^v),\\
    \mathcal{F}_L^v &= \text{Conv}(\text{Concat}(I^v, R^v)),
    \label{eq:encoder}
\end{align}
where $\mathcal{F}_H^v$ and $\mathcal{F}_L^v$ represent the high and low-level feature maps of image $I^v$, respectively.

\textbf{Geometry-aware Transformer}\quad The geometry-aware transformer aims to efficiently lift image features to 3D. The proposal transformer and reconstruction transformer are both instances of this architecture. Previous methods~\cite{hong2023lrm, tang2024lgm, xu2024grm, wei2024meshlrm, zhang2024gslrm} use tri-planes or pixel-aligned Gaussians to represent 3D contents. However, these data structures make it hard to utilize the projection relationships, causing dense computations. Instead, we use 3D anchor points, which serve as proxies for their surrounding points, significantly reducing the number of points we need to process. As detailed in Figure~\ref{fig:pipeline}, each transformer block contains a self-attention layer, a deformable cross-attention layer and a feed-forward network (FFN). The model takes $N$ anchor point features $\mathcal{F}_A = \{\boldsymbol{f}_i\}_{i=1}^N$ as input tokens. Each token $\boldsymbol{f}_i$ comprises the coordinate of the corresponding point and a shared learnable feature.

For the self-attention layer, a crucial problem is how to inject positional information into the sparse 3D tokens. We extend the Rotary Positional Embedding (RoPE)~\cite{rope} to 3D conditions for relative positional embedding. For a query $\boldsymbol{q}_{\boldsymbol{m}}$ and a key $\boldsymbol{k}_{\boldsymbol{n}}$ at absolute position $\boldsymbol{m}$ and $\boldsymbol{n}$, we ensure that the inner product of embedded values reflects only the relative position information $\boldsymbol{m} - \boldsymbol{n}$. A direct yet promising way is splitting the features into three parts and applying RoPE~\cite{rope} on each part with x, y, and z positions respectively.

As we can locate each anchor point in the 3D space, a possible way to lift 2D features to 3D is to project them to the feature maps with known poses and average the corresponding features. However, this method assumes an accurate anchor position, an equal contribution of all images and a good 3D correspondence of input images, which is often impractical, especially in 3D generation tasks. To tackle these issues, we employ deformable attention~\cite{deform_detr, bevformer, surroundocc} for a robust fusion of image features. Given a 3D anchor point feature $\boldsymbol{f}_i$, its spatial coordinate $\boldsymbol{x}_i$ and multiple feature maps $\{\mathcal{F}^v\}_{v=1}^V$, the deformable attention mechanism is formulated as:
\begin{equation}
    \text{DeformAttn}(\boldsymbol{f}_i, \boldsymbol{x}_i, \{\mathcal{F}^v\}_{v=1}^V) = 
    \sum_{v=1}^V w_{v} [
    \sum_{k=1}^K A_{k} \mathcal{F}^v \left<\boldsymbol{p}_{iv} + \Delta \boldsymbol{p}_{ivk} \right>
    ],
    \label{eq:deform_attn}
\end{equation}
where $k$ indexes the sampled keys and $K$ is the total sampled key numbers. $\boldsymbol{p}_{iv}$ is the projected 2D coordinate on feature map $\mathcal{F}^v$ and $\Delta \boldsymbol{p}_{ivk}$ is the sampled offset. $\left<\cdot\right>$ indicates the interpolation operation. $A_k$ is the attention weight predicted from $\boldsymbol{f}_i$. $w_v$ is a per-view weight derived from the feature it weights. Notably, the prediction of $\Delta \boldsymbol{p}_{ivk}$ allows the network to correct the geometry error of anchor points and the inconsistency of input images; The $w_v$ enables different importance levels for each image. To further enhance the representation ability of the model, this mechanism is extended to multi-head and multi-scale conditions.

Given input tokens $\mathcal{F}_A^{in}$, the transformer block enhances these tokens through a series of sophisticated transformations described as follows:
\begin{align}
    \mathcal{F}_A^{self} &= \mathcal{F}_A^{in} + \text{SelfAttn}(\text{RMSNorm}(\mathcal{F}_A^{in})),\\
    \mathcal{F}_A^{cross} &= \mathcal{F}_A^{self} + \text{DeformCrossAttn}(\text{RMSNorm}(\mathcal{F}_A^{self}), \{(\mathcal{F}_H^v, \mathcal{F}_L^v)\}_{v=1}^V),\\
    \mathcal{F}_A^{out} &= \mathcal{F}_A^{cross} + \text{FFN}(\text{RMSNorm}(\mathcal{F}_A^{cross})).
\label{eq:decoder}
\end{align}
This design introduces several improvements over the original transformer architecture~\cite{transformer}. By incorporating RMSNorm~\cite{rmsnorm} for normalization and SiLU~\cite{silu} for activation, we achieve more stable training dynamics and better performance.

\textbf{Post-processing}\quad The proposal network takes a low-resolution dense grid ($16^3$) as anchor points. The output is upsampled to a high-resolution grid ($128^3$) with a linear layer. This grid is formulated to represent the occupancy probability of the corresponding area ($[-0.5, 0.5]^3$). The reconstruction transformer takes occupied voxels as anchor points. Each output token $\boldsymbol{f}_i$ is decoded into multiple 3D Gaussians $\{\boldsymbol{G}_{ij}\}_{j=1}^{M}$ with a multilayer perceptron. The 3D Gaussian $\boldsymbol{G}_{ij}$ is parameterized by the offset $\boldsymbol{o}_{ij}$ regarding the anchor points, 3-channel RGB $\boldsymbol{c}_{ij}$, 3-channel scale $\boldsymbol{s}_{ij}$, 4-channel rotation quaternion $\boldsymbol{\sigma}_{ij}$, and 1-channel opacity $\alpha_{ij}$.
We employ activation functions to limit the range of the offset, scale and opacity for better training stability similar to~\cite{tang2024lgm}:
\begin{align}
    \boldsymbol{o}_{ij} &= \text{Sigmoid}(\boldsymbol{o}_{ij}') \cdot o_{\rm max},\\
    \boldsymbol{s}_{ij} &= \text{Sigmoid}(\boldsymbol{s}_{ij}') \cdot s_{\rm max} ,\\
    \alpha_{ij} &= \text{Sigmoid}(\alpha_{ij}'),
\label{eq:decode_gs}
\end{align}
where $o_{\rm max}, s_{\rm max}$ are predefined maximum values of offsets and scales. Given target camera views $\{\boldsymbol{c}_{t}\}_{t=1}^{T}$, the 3D Gaussians can be further rendered into images $\{\hat{I}_{t}\}_{t=1}^{T}$, alpha masks $\{\hat{M}_{t}\}_{t=1}^{T}$ and depth maps $\{\hat{D}_{t}\}_{t=1}^{T}$ through Gaussian splatting~\cite{3dgs}.

\subsection{Training Objectives}
\label{sec:method_to}

We employ a two-stage training mechanism for our model. In the first stage, we train the proposal transformer using 3D occupancy ground truth. This stage presents a challenge as it involves a highly unbalanced binary classification task; only about 5\% of the voxels are occupied. To address this imbalance, we employ a combination of binary cross-entropy loss and the scene-class affinity loss, as proposed in~\cite{monoscene}, to supervise the training process. For the generation of ground truth data, see~\ref{sec:occ}.

For the second stage, we supervise the rendered $T$ images, alpha masks and depth maps with corresponding ground truth:
\begin{align}
    &\mathcal{L} = \sum_{t=1}^{T} \left( \mathcal{L}_{\rm img}(\hat{I}_t, I_t) + \mathcal{L}_{\rm mask}(\hat{M}_t, M_t) + 0.2\mathcal{L}_{\rm depth}(\hat{D}_t, D_t, I_t) \right),\\
    &\mathcal{L}_{\rm img}(\hat{I}_t, I_t) = ||\hat{I}_t - I_t||_2 + 2 \mathcal{L}_{\rm LPIPS}(\hat{I}_t, I_t),\\
    &\mathcal{L}_{\rm mask}(\hat{M}_t, M_t) = ||\hat{M}_t - M_t||_2,\\
    &\mathcal{L}_{\rm depth}(\hat{D}_t, D_t, I_t) = \frac{1}{|\hat{D}_t|} \left|\left| \exp(-\Delta I_t) \odot \log(1+|\hat{D}_t - D_t|) \right|\right|_{1},
\label{eq:loss}
\end{align}
where $\mathcal{L}_{\rm LPIPS}$ is the perceptual image patch similarity loss~\cite{lpips}, $|\hat{D}_t|$ is the total number of pixels in $|\hat{D}_t|$, $\Delta I_t$ is the gradient of the current RGB image and $\odot$ is the element-wise multiplication operation. As demonstrated in~\cite{dnsplatter}, applying a logarithmic penalty and weighting the per-pixel depth errors with the image gradients result in a smoother geometric representation.

\section{Experiments}

\subsection{Datasets}

\textbf{G-buffer Objaverse (GObjaverse)~\cite{richdreamer}:} Used for training. Derived from the original Objaverse~\cite{objaverse} dataset, GObjaverse includes high-quality renderings of albedo, RGB, depth, and normal images. These images are generated through a hybrid technique combining rasterization and path tracing. The dataset comprises approximately 280,000 normalized 3D models scaled to fit within a cubic space of $[-0.5, 0.5]^3$. GObjaverse employs a diverse camera setup involving:
\begin{itemize}
    \item Two orbital paths yielding 36 views per model. This includes 24 views at elevations between 5° and 30° (incremented by 15° rotations) and 12 views at near-horizontal elevations from -5° to 5° (with 30° rotation steps).
    \item Additional top and bottom views for comprehensive spatial coverage.
\end{itemize}

\textbf{Google Scanned Objects (GSO)~\cite{gso}:} Used for evaluation, this dataset is rendered similarly to GObjaverse. A random subset of 100 objects is selected to streamline the evaluation process.

\textbf{OmniObject3D~\cite{omniobject3d}:} Also used for evaluation, this dataset is consistently rendered like GObjaverse. A random subset of 100 objects is chosen for efficient evaluation.

\subsection{Implementation details}
Our model features 330 million parameters distributed across two distinct image encoders and two transformers. The first encoder processes geometry with the 6-layer proposal transformer, while the second focuses more on textures crucial with the 16-layer reconstruction transformer. During training, we maintain a maximum number of transformer input tokens of 4k and randomly select 8 views from a possible 38 for supervision. From these 8 views, we randomly select 1 to 7 views as inputs to predict the remaining views. This flexibility in view selection not only tests the robustness of our method but also mimics real-world scenarios where complete data may not always be available. Both input and rendering resolutions are maintained at 448x448 pixels. At the testing and inference stages, we use a resolution of 512x512 to align with existing methods. Besides, the number of input tokens is extended to 16k during testing, showcasing its scalability without the need for fine-tuning. Detailed information on our model's architecture and training procedures can be found in Section~\ref{supp:hyper}.

\subsection{Quantitative Results}

\begin{table}[t]
  \renewcommand{\tabcolsep}{1.5mm}
  \caption{Quantitative results on Google Scanned Objects (GSO)~\cite{gso}, where we used six views for inputs and four for evaluation. Inference time and memory usage account only for the reconstruction process. \textbf{Bold} and \underline{underline} denote the highest and second-highest scores, respectively.}
  \centering
  \begin{tabular}{@{}l|c|c|c|c|c|c|c}
    \toprule
        Method & PSNR $\uparrow$ & SSIM $\uparrow$ & LPIPS $\downarrow$ & CD $\downarrow$ & FS $\uparrow$ & Inf. Time (s) & Memory (GB)\\
        \midrule
        LGM &20.76&0.832&0.227&0.295&0.703&\textbf{0.07}&\text{7.23}\\
        CRM &22.78&0.843&0.190&0.213&0.831&\underline{0.30}&\underline{5.93}\\
        InstantMesh &\underline{23.19}&\underline{0.856}&\textbf{0.166}&\underline{0.186}&\underline{0.854}&\text{0.78}&\text{23.12}\\
        Ours &\textbf{23.57}&\textbf{0.872}&\underline{0.167}&\textbf{0.167}&\textbf{0.892}&0.67&\textbf{4.92}\\
    \bottomrule
  \end{tabular}
  \label{tab:gso}
\end{table}

\begin{table}[t]
  \renewcommand{\tabcolsep}{1.5mm}
  \caption{Quantitative results on OmniObject3D~\cite{omniobject3d}. \textbf{Bold} and \underline{underline} denote the highest and second-highest scores, respectively.}
  \centering
  \begin{tabular}{@{}l|c|c|c|c|c}
    \toprule
        Method & PSNR $\uparrow$ & SSIM $\uparrow$ & LPIPS $\downarrow$ & CD $\downarrow$ & FS $\uparrow$\\
        \midrule
        LGM & 21.94 & 0.824 & 0.203 & 0.256 & 0.787\\
        CRM & 23.12 & 0.855 & 0.175 & 0.204 & 0.810\\
        InstantMesh &\underline{23.86} & \underline{0.860} & \underline{0.139} & \underline{0.178} & \underline{0.834}\\
        Ours & \textbf{24.74} & \textbf{0.883} & \textbf{0.134} & \textbf{0.156} & \textbf{0.863}\\
    \bottomrule
  \end{tabular}
  \label{tab:gso}
\end{table}

\begin{table}[t]
\footnotesize
\centering
\caption{Quantitative results on Google Scanned Objects (GSO) with different numbers of input views. We keep the same four views for testing while changing the number of input views. \textbf{Bold} denotes the highest score.}
\begin{adjustbox}{center}
    \begin{tabular}{c|cc|cc|cc|cc}
    \toprule
    \multirow{2}{*}{Num Input} & \multicolumn{2}{c|}{PSNR} & \multicolumn{2}{c|}{SSIM} & \multicolumn{2}{c|}{Inf. Time (s)} & \multicolumn{2}{c}{Memory (GB)} \\
     & \text{InstantMesh} & \text{Ours}  & \text{InstantMesh} & \text{Ours}  & \text{InstantMesh} & \text{Ours}  & \text{InstantMesh} & \text{Ours} \\
    \midrule
        4      & \textbf{22.87} & 22.84 & 0.832 & \textbf{0.851} & \text{0.68} & \textbf{0.51} & \text{22.09} & \textbf{4.30} \\
        8      & 23.22 & \textbf{23.82} & 0.861 & \textbf{0.883} & \text{0.87} & \textbf{0.84} & \text{24.35} & \textbf{5.50} \\
        12    & 23.05 & \textbf{24.43} & 0.843 & \textbf{0.892} & \textbf{1.07} & \text{1.16} & \text{24.62} & \textbf{6.96} \\
        16    & 23.15 & \textbf{24.79} & 0.861 & \textbf{0.903} & \textbf{1.30} & \text{1.51} & \text{26.69} & \textbf{8.23} \\
        20    & 23.25 & \textbf{25.13} & 0.895 & \textbf{0.905} & \textbf{1.62} & \text{1.84} & \text{28.73} & \textbf{9.43} \\
    \bottomrule
    \end{tabular}
\end{adjustbox}
\label{tab:num_view}
\end{table}

We evaluated the quality of reconstructed assets from sparse view inputs by analyzing both 2D visual and 3D geometric aspects on the GSO and OmniObject3D dataset~\cite{gso}. Visual quality was assessed by comparing rendered views to ground truth images using metrics such as PSNR, SSIM, and LPIPS. Geometric accuracy was evaluated by aligning our models to the ground truth coordinate systems and measuring discrepancies using Chamfer Distance and F-Score at a threshold of 0.2, with point samples totalling 16,000 from the ground truth surfaces. Our method was quantitatively compared against established baselines, including LGM~\cite{tang2024lgm}, CRM~\cite{wang2024crm}, and InstantMesh~\cite{xu2024instantmesh}. We avoided comparisons with proprietary methods due to the unavailability of their test splits. Similarly, we excluded comparisons with OpenLRM~\cite{openlrm} and TripoSR~\cite{tochilkin2024triposr} as these methods are tailored for single image inputs, which would be unfair to compare with.

Our approach achieved state-of-the-art performance in four out of the five metrics studied. Although InstantMesh showed slightly higher LPIPS on the GSO dataset, attributed to its mesh-based smoothing capabilities, our method demonstrated superior geometric accuracy, benefiting from explicit modelling of the 3D-to-2D relationship.

In another experiment, outlined in Table~\ref{tab:num_view}, we observed a notable trend: our model's performance consistently improves with more input views while maintaining low computational costs. This indicates robust scalability, a critical feature for practical applications. In contrast, the performance of InstantMesh~\cite{xu2024instantmesh}, does not follow this pattern. Specifically, InstantMesh shows a decline in performance when the input views increase to 12. This degradation could be due to two primary factors. First, the low-resolution tri-planes may reach their maximum capacity to represent details. Second, the model tends to oversmooth details when handling a large volume of image tokens. Our approach strategically addresses these issues. We employ an extendable sequence of 3D tokens that can be dynamically adjusted to fit the resolution requirements. Additionally, our model features deformable attention mechanisms that intelligently focus on the most pertinent information, preventing the loss of critical details.

\begin{figure}[t]
  \centering
  \includegraphics[width=1.0\textwidth]{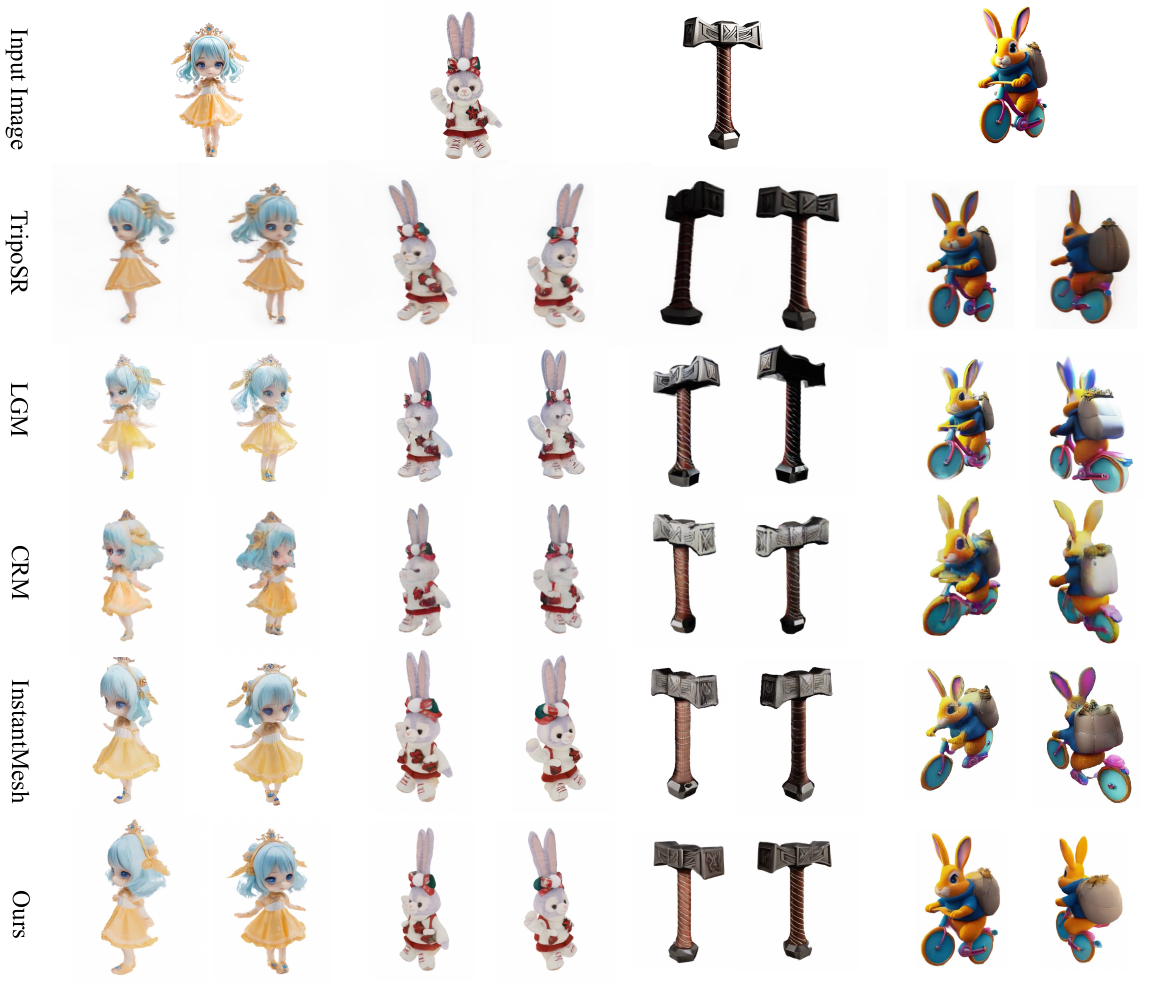}
  \caption{Qualitative comparisons of different image-3D methods. \textbf{Better viewed when zoomed in.}}
  \label{fig:qua_results}
\end{figure}

\begin{figure}[h]
  \centering
  \includegraphics[width=1.0\textwidth]{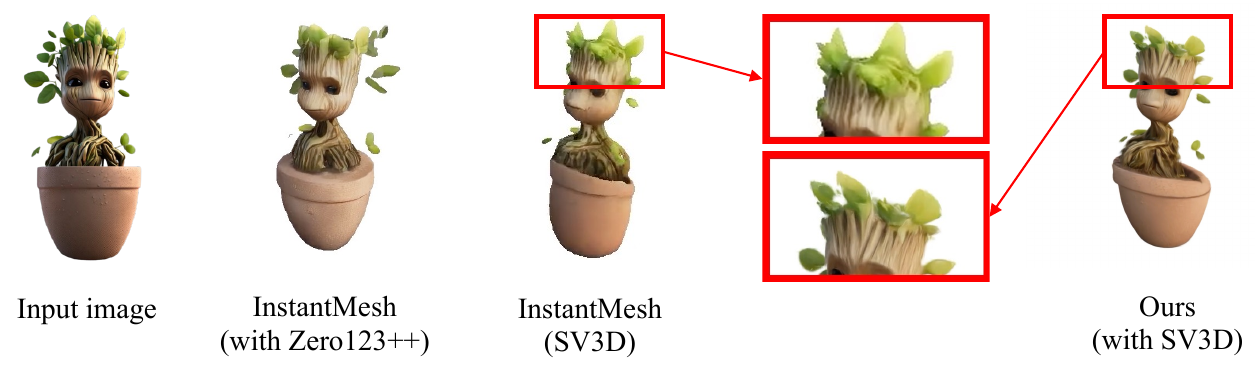}
  \caption{Qualitative comparison concerning scalability in input views.}
  \label{fig:qua_sv3d}
\end{figure}

\subsection{Qualitative Results}

We conducted a qualitative analysis comparing our method with several LRM-based baselines, including TripoSR \cite{openlrm}, LGM \cite{tang2024lgm}, CRM \cite{wang2024crm}, and InstantMesh \cite{xu2024instantmesh}, maintaining their original settings to ensure optimal performance. In our approach, we utilized the SV3D~\cite{sv3d} technology to generate 21 multi-view images, significantly enhancing the resolution and textural details of the 3D Gaussians produced, as illustrated in Figure~\ref{fig:qua_results}. Furthermore, as shown in Figure~\ref{fig:qua_sv3d}, employing InstantMesh to reconstruct these images did not yield satisfactory outcomes, corroborating our quantitative findings. This demonstrates the superior capability of our method in handling more complex 3D reconstructions.

\subsection{Ablation Study}
In this part, We provide ablation studies for the key designs of our method as shown in Table~\ref{tab:abla}. Due to the limited computational sources, the ablation is done using a smaller reconstruction model (12 layers) and lower resolution (224x224).

\begin{table}[h]
\caption{Ablation study of some key designs. Models are tested on the GSO dataset~\cite{gso}. Upper: 6 input views and 4 testing views. Lower: different input views. \textbf{Bold} and \underline{underline} denote the highest and second-highest scores, respectively.}
\centering
  \begin{tabular}{@{}l|c|c|c}
    \toprule
     Method & PSNR $\uparrow$ & SSIM $\uparrow$ & LPIPS $\downarrow$ \\
    \midrule
    W/o Plücker rays & 20.64 & 0.826 & 0.244 \\
    W/o low-level features & 20.29 & 0.817 & 0.246 \\
    W/o high-level features & 15.85 & 0.798 & 0.289 \\
    W/o 3D RoPE & 20.52 & 0.827 & 0.224 \\
    Fixed \# input views & \textbf{20.97} & \textbf{0.839} & \underline{0.220} \\
    Full model & \underline{20.73} & \underline{0.831} & \textbf{0.216} \\
    \bottomrule
  \end{tabular}
  
    \vspace{8pt}
    \begin{tabular}{l|cc|cc|cc}
    \toprule
        & \multicolumn{2}{c|}{4 Inputs} &  \multicolumn{2}{c|}{8 Inputs} &  \multicolumn{2}{c}{12 Inputs} \\ 
        Method & PSNR \(\uparrow\) & SSIM \(\uparrow\)  & PSNR \(\uparrow\) & SSIM \(\uparrow\) & PSNR \(\uparrow\) & SSIM \(\uparrow\) \\ 
        \midrule
        Fixed \# input views & \underline{19.72} & \underline{0.822} & \underline{20.85} & \underline{0.833} & \underline{21.43} & \underline{0.838} \\
        Full model & \textbf{19.94} & \textbf{0.835} & \textbf{21.16} & \textbf{0.840} & \textbf{22.04} & \textbf{0.853} \\
    \bottomrule
    \end{tabular}
    
  \label{tab:abla}
\end{table}

\begin{figure}[h]
  \centering
  \includegraphics[width=0.9\textwidth]{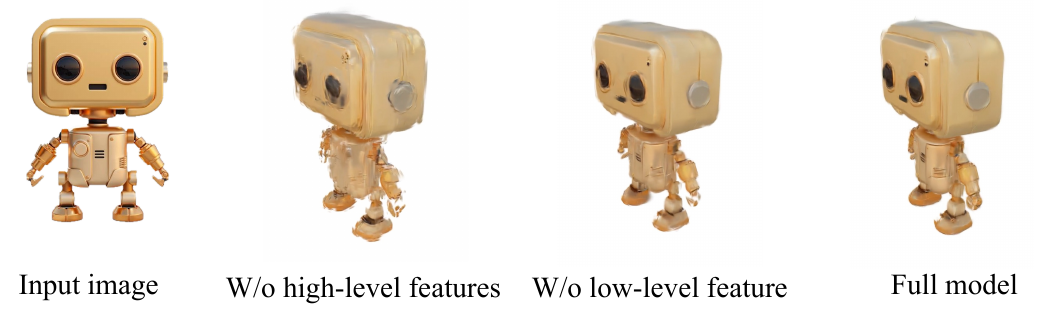}
  \caption{Effects of excluding high-level and low-level features in the image encoder.}
  \label{fig:abla}
\end{figure}

\textbf{Hierarchical Image Encoder}\quad Our ablation study underscores the critical role of hierarchical image features in reconstruction tasks, which necessitate both high-level semantic information (e.g., object identity and arrangement) and low-level texture information (e.g., surface patterns and colors). As illustrated in Figure~\ref{fig:abla}, the absence of high-level features leads to model instability, while omitting low-level features results in a loss of textural detail. This dual requirement emphasizes the model's reliance on a comprehensive feature set for accurate image reconstruction. We also performed an ablation study regarding the Plücker ray embeddings in the low-level encoder. These coordinates assist the model in learning camera directions, contributing to an improved performance.

\textbf{3D RoPE}\quad In transformer-based architectures, the role of positional embeddings is critical for accurately interpreting sequence data positions. A key question arises: With the reconstruction transformer employing deformable cross-attention to elevate 2D features to 3D, is positional embedding still necessary? Our ablation studies confirm its necessity. Notably, 3D RoPE significantly enhances the model's ability to handle longer sequences. For instance, increasing the sequence length from 4k to 16k elements, models equipped with 3D RoPE exhibited a PSNR improvement of 0.4, compared to a 0.2 improvement in models lacking 3D RoPE. This observation aligns with the 1D RoPE~\cite{rope}.

\textbf{Dynamic Input}\quad The ablation study demonstrates a decrease in performance when employing our dynamic input view strategy compared to the fixed 6-input view setting when the training and testing phases were consistent. Despite this, the dynamic input strategy enhances the model's ability to generalize across different input configurations. This adaptability is critical for handling more complex scenarios, aligning with our primary objectives.

\textbf{Deformable attention}\quad As shown in Table~\ref{tab:abla_deform}, the ablation results indicate that increasing the number of sampling points in the deformable attention generally improves performance. Given the trade-off between computational cost and performance gain, we find that using 8 sampling points strikes the best balance.

\begin{table}[th]
  \caption{Ablation study of deformable attention. `0 sampling points' means directly using the projected points without any deformation. \textbf{Bold} and \underline{underline} denote the highest and second-highest scores, respectively.}
  \centering
  \begin{tabular}{@{}l|c|c|c}
    \toprule
     Method & PSNR $\uparrow$ & SSIM $\uparrow$ & LPIPS $\downarrow$ \\
    \midrule
    \text{0 sampling points} & 19.52 & 0.802 & 0.265 \\
    \text{4 sampling points}  & 20.21 & 0.819 & 0.238 \\
    \text{8 sampling points}  & \underline{20.73} & \underline{0.839} & \underline{0.220} \\
    \text{16 sampling points} & \textbf{20.80} & \textbf{0.846} & \textbf{0.219} \\
    \bottomrule
  \end{tabular}
  \label{tab:abla_deform}
\end{table}

\section{Conclusion}
In this paper, we present GeoLRM, a geometry-aware large reconstruction model designed to improve the efficiency and quality of 3D generation. Our approach distinguishes itself from previous methods by effectively utilizing the inherent sparsity of 3D structures and explicitly integrating geometric relationships between 3D and 2D images. The GeoLRM framework employs a 3D-aware transformer architecture that predicts 3D Gaussians through a sophisticated coarse-to-fine methodology. Initially, a proposal network estimates coarse occupancy grids, which serve as foundational 3D anchor points for subsequent refinement. The second stage leverages deformable cross-attention to enhance the 3D structure, integrating detailed textural information. Extensive experiments validate that GeoLRM can process higher resolutions and accommodate denser image inputs, outperforming existing models in terms of detail and accuracy. This innovation demonstrates significant potential for real-world applications, particularly in domains where dense view inputs can enhance output quality and user experience. GeoLRM's ability to handle up to 21 images efficiently underscores its scalability and adaptability, paving the way for integration with advanced video generation technologies.

\section{Limitation}
While GeoLRM achieves impressive reconstruction quality, it does so through a two-stage process, which is not inherently end-to-end. This segmentation can lead to the accumulation of errors. The reliance on a proposal network is currently indispensable due to the computational intensity of processing Gaussian points across the entire 3D space. This necessity introduces potential inefficiencies and constraints that could hinder real-time applications. Future research will focus on developing an end-to-end solution that integrates these stages seamlessly, reducing error propagation and optimizing processing time. By addressing these limitations, we aim to enhance the model's robustness and applicability across a broader range of 3D generation tasks.

\section*{Acknowledgement}
This work was supported in part by the Beijing Natural Science Foundation under Grant No. L247009 and in part by Young Elite Scientists Sponsorship Program by CAST (No. 2024QNRC003).

{
    \small
    \bibliographystyle{plain}
    \bibliography{geolrm}
}

\clearpage
\appendix
\section{Appendix}

\renewcommand\thetable{\Alph{table}}
\renewcommand\thefigure{\Alph{figure}}
\setcounter{table}{0}
\setcounter{figure}{0}

\subsection{Occupancy Ground Truth}
\label{sec:occ}

Previous studies~\cite{renderocc, occnerf, selfocc} have investigated the task of vision-centric occupancy prediction. However, these approaches often exhibit significant performance discrepancies when compared to 3D methods. To bridge this gap, we leverage depth maps from the GObjaverse dataset to generate accurate 3D occupancy ground truths. This process begins by transforming each pixel in the depth map, represented as $\mathbf{p^i} = [u, v, 1]^T$, into a point in world coordinates. This transformation uses both the intrinsic matrix $K$ and the extrinsic parameters $T$, consisting of a rotation matrix $R$ and a translation vector $\mathbf{t}$, as shown in the equation:
\begin{align}
    \mathbf{p^w} = R(d\cdot K^{-1} \mathbf{p^i}) + \mathbf{t},
\end{align}
where $d$ denotes the depth at pixel $\mathbf{p^i}$. Subsequently, these world coordinates are voxelized to pinpoint occupied voxel centres:
\begin{align}
    V = \left\{\left\lfloor\frac{P}{\epsilon}\right\rceil\right\} \cdot \epsilon,
\end{align}
where $P$ includes all points in three-dimensional space, $V$ represents the voxel centers, and the voxel size $\epsilon$ is set at $1/128$. The voxelization helps in reducing redundancy by removing duplicate entries.

\begin{figure}[th]
  \centering
  \includegraphics[width=1.0\textwidth]{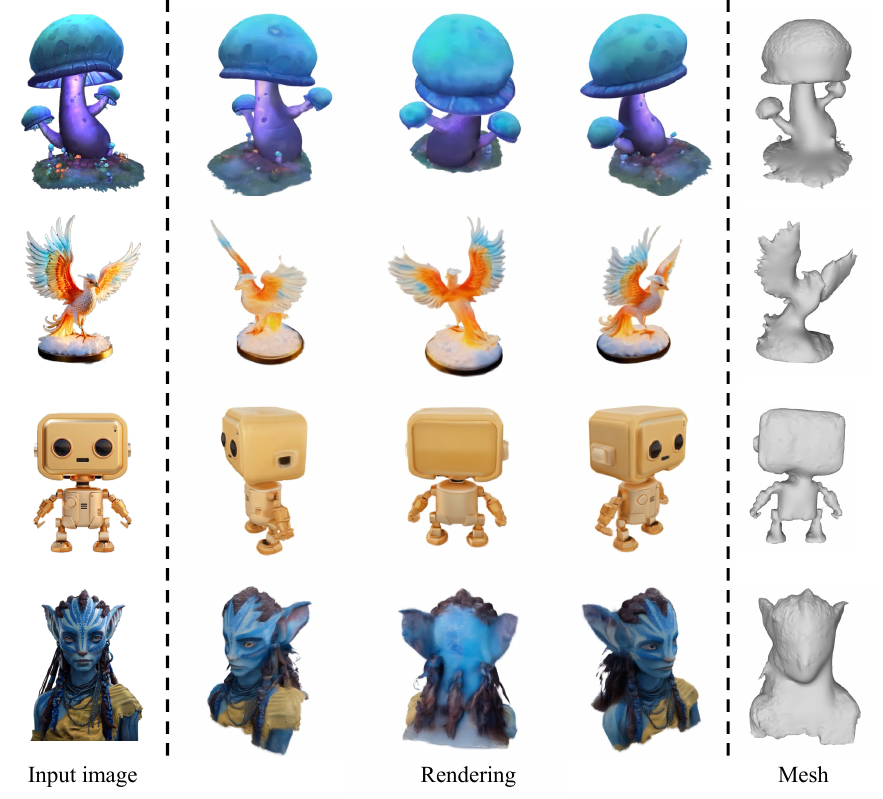}
  \caption{Image-to-3D generation with mesh extraction results.}
  \label{fig:mesh}
\end{figure}

\subsection{Mesh Extraction from 3D Gaussians}
\begin{figure}[htb]
  \centering
  \includegraphics[width=1.0\textwidth]{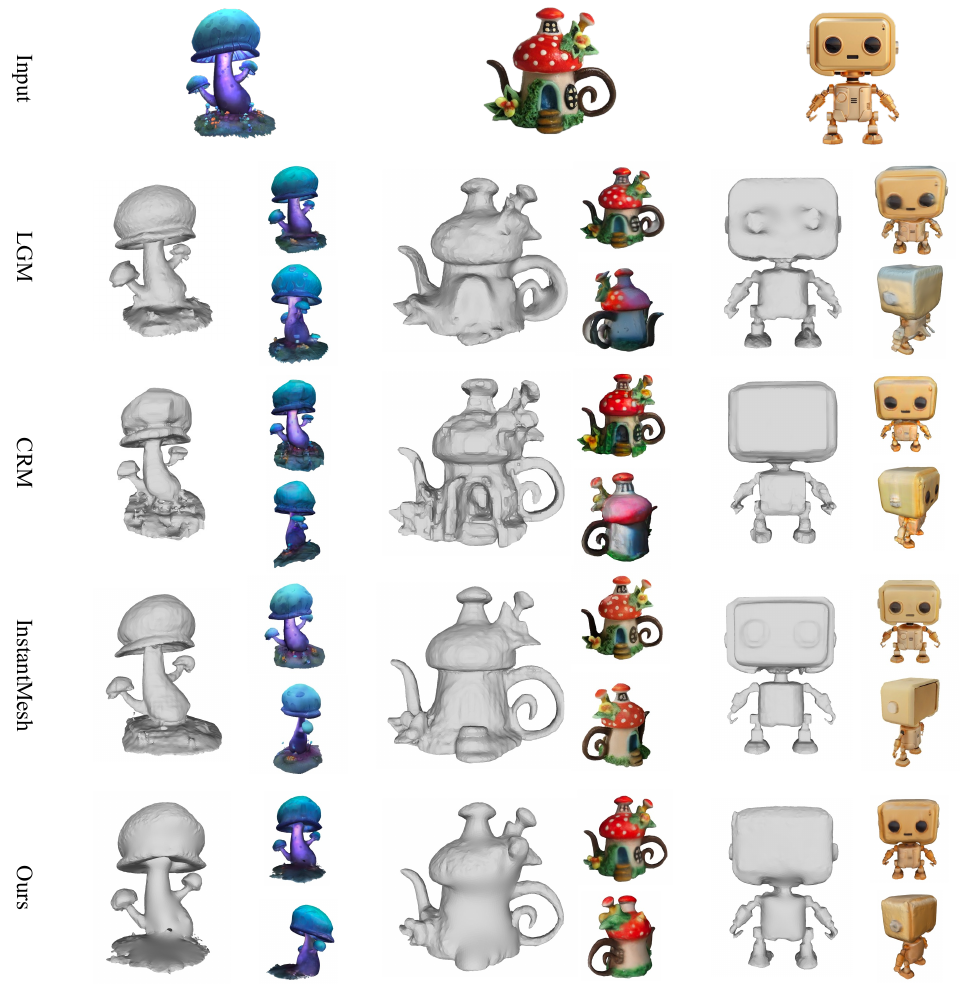}
  \caption{Comparison of the generated meshes.}
  \label{fig:mesh_comp}
\end{figure}

We adopt the mesh extraction pipeline from~\cite{tang2024lgm} to derive high-quality mesh representations from 3D Gaussians. Figure~\ref{fig:mesh} illustrates the mesh generation results of our method, while Figure~\ref{fig:mesh_comp} compares our generated mesh with other techniques. The results demonstrate the effectiveness of our approach, despite some loss of detail during conversion.

\subsection{More Implementation Details}
\label{supp:hyper}

We illustrate the details of network architecture and training procedure in Table~\ref{tab:hyper}. We train both the proposal transformer and the reconstruction transformer for 12 epochs on GObjaverse~\cite{richdreamer}, which takes 0.5 and 2 days respectively on 32 A100 40G. For the proposal transformer, we use a batch size of 2 per GPU and apply mixed-precision training with BF16 data type. For the reconstruction transformer, we use a batch size of 1 per GPU and keep the full precision. We note that the second stage is particularly sensitive to the data type and would fail if using mixed-precision.

\begin{table}[th]
\centering
\caption{Implementation details.}
\begin{tabular}{l|l|l}
\toprule

\multirow{4}{*}{Proposal Transformer}   
                           & Image encoder              &  DINOv2 (ViT-B/14) + Conv  \\
                           & \# layers                  &  6  \\
                           & \# attention head          & 16  \\
                           & \# deformed points         & 8   \\
                           & Image feature dimension    & 384  \\
                           & 3D feature dimension       & 384  \\
                           & Max sequence length        & 4096 \\

\midrule
\multirow{5}{*}{Reconstruction Transformer}
                           & Image encoder              &  DINOv2 (ViT-B/14) + Conv  \\
                           & \# layers                  & 16  \\
                           & \# attention head          & 16  \\
                           & \# deformed points         & 8   \\
                           & Image feature dimension    & 384  \\
                           & 3D feature dimension       & 768  \\
                           & Max sequence length        & 4096 \\
                           & \# Gaussians per token     & 32   \\
\midrule
\multirow{8}{*}{Training details}
                           & Epoch                      &    12    \\
                           & Learning rate              &    1e-4    \\
                           & Learning rate scheduler    &    Cosine \\
                           & Optimizer                  &    AdamW       \\
                           & (Beta1, Beta2)             & (0.9, 0.95) \\
                           & Weight decay               &     0.05      \\
                           & Warm-up                    &      1500       \\
                           & Gradient accumulation      & 8 \\
                           & Gradient clip              & 4 \\
                           & \# GPU                     & 32 \\

\bottomrule
\end{tabular}
\label{tab:hyper}
\end{table}

\subsection{Social Impact}
3D AIGC is transforming sectors by automating realistic 3D model creation. In entertainment, it streamlines film and game production, reducing costs and enhancing experiences. Education benefits from immersive VR simulations for deeper learning. Architecture sees rapid design visualization and urban planning improvements. Challenges include job displacement and ethical concerns over content authenticity. Addressing these requires legal and policy measures, such as clear copyright laws and standards to protect intellectual property. Developing advanced content moderation tools can detect false content, and enhancing AI security can prevent misuse. By focusing on these solutions, we can mitigate negative impacts and maximize the positive contributions of 3D AIGC to society.

\end{document}